\documentclass[11pt]{article}

\usepackage{acl}
\usepackage{svg}
\usepackage{times}
\usepackage{latexsym}

\usepackage{tabularx}
\usepackage{booktabs}

\usepackage[T1]{fontenc}
\usepackage[utf8]{inputenc}

\usepackage{microtype}

\title{There is No Big Brother or Small Brother: Knowledge Infusion in Language Models for Link Prediction and Question Answering}

\author{Ankush Agarwal$^{1*}$, Sakharam Gawade$^{1*}$, \\\textbf{Sachin Channabasavarajendra$^2$, Pushpak Bhattacharyya$^1$}\\
        $^1$IIT Bombay, \\$^2$Honeywell Technology Solutions Pvt Ltd \\
        \texttt{\{ankushagrawal, sakharamg, pb\}@cse.iitb.ac.in,}\\
         \texttt{sachin.channabasavarajendra@honeywell.com}}
\begin{document}

\maketitle
\def\thefootnote{*}\footnotetext{Equal contribution}\def\thefootnote{\arabic{footnote}}

\begin{abstract}

The integration of knowledge graphs with deep learning is thriving in improving the performance of various natural language processing (NLP) tasks. In this paper, we focus on knowledge-infused link prediction and question answering using language models, T5, and BLOOM across three domains: Aviation, Movie, and Web. 
In this context, we infuse knowledge in large and small language models and study their performance, and find the performance to be similar. For the link prediction task on the Aviation Knowledge Graph, we obtain a 0.2 hits@1 score using T5-small, T5-base, T5-large, and BLOOM. Using template-based scripts, we create a set of 1 million synthetic factoid QA pairs in the aviation domain from National Transportation Safety Board (NTSB) reports. On our curated QA pairs, the three models of T5 achieve a 0.7 hits@1 score. 
We validate our findings with the paired student t-test and Cohen's kappa scores. 
For link prediction on Aviation Knowledge Graph using T5-small and T5-large, we obtain a Cohen's kappa score of 0.76, showing substantial agreement between the models. Thus, we infer that small language models perform similar to large language models with the infusion of knowledge.
% I have updated the abstract, and added kohen's kappa score
\end{abstract}

\section{Introduction} \label{intro}

A large number of pre-trained language models (LMs) are used for downstream tasks, such as Question Answering (QA). Generally, these language models are trained on generic domain data, such as Web data and News Forums. 
% Recently, work with LMs is also focused on domain-related fields, namely, healthcare, radiology, and aviation. 
Recently, LMs are used for downstream tasks in domain-specific fields, namely, healthcare  \cite{michalopoulos-etal-2021-umlsbert}, radiology \cite{kaveriradiology}, and aviation \cite{agarwal2022knowledge}.
% For tasks such as Information Extraction (IE) and QA, it is observed that Knowledge graphs (KGs) play a vital role in improving performance by infusing knowledge. 
For tasks such as Information Extraction (IE) and Question Answering (QA), Knowledge Graphs (KGs) are used as a source of external knowledge to boost the performance of models.
To a great extent, researchers focus on the synergy of Knowledge Graph and Deep Learning \cite{kvmNet,embedkgqa,saxena2022sequence}.
% need to cite more papers to show that there is focus on the synergy, e.g. EmbedKGQA, we can check survey papers on KGDL to get more papers
With the increase in data, it is observed that larger models are preferred for different tasks across various domains. 
% The pre-trained language models are trained on a large amount of data and have millions of parameters.
\par
The Large Language Models (LLMs) are preferred to obtain better results than small or non-pre-trained models as they have a vast number of parameters and have been trained on a large amount of data.
%size of model 
But, the larger model increases the need for computation power and training time. In this paper, we show that small and large models perform likewise with the infusion of knowledge. We can use non-pre-trained models for different tasks across domains that require less computation power and time and still attain the same performance as pre-trained models.  
\par
% Need to introduce AviationQA before mentioning it
We validate our hypothesis with the LLMs, \textit{i.e.}, T5 \cite{raffel2020exploring} \& BLOOM\footnote{\url{https://huggingface.co/bigscience/bloom}}. We perform two tasks: a) Link Prediction, and b) Question Answering on different datasets: a) Aviation Knowledge Graph (AviationKG) \cite{agarwal2022knowledge}, and Aviation QA pairs (section \ref{aviationQA}), b) Movie Knowledge Base (MovieKB) \& MetaQA (a set of QA pairs), both present in the MetaQA dataset \cite{zhang2018variational},
% Meta KB is known name?
and c) Complex Web Questions (CWQ) \cite{talmor2018web}, which uses subsets of Freebase \cite{chah2017freebase}. We perform hypothesis testing to validate our hypothesis. We use paired Student T-test and attempt to reject our hypothesis that models have a negligible difference in performance. But, we were not able to repudiate our hypothesis. To strengthen our findings, we use Cohen's kappa measure and show significant agreement between models.  
\par
Our \textbf{contributions} are as follows:
\begin{enumerate}

    \item We create a synthetic dataset, AviationQA \footnote{\url{https://github.com/ankush9812/Aviation-Question-Answer-Pairs}}, a set of 1 million factoid QA pairs from 12,000 National Transportation Safety Board (NTSB) reports using templates explained in section \ref{aviationQA}. These QA pairs contain questions such that answers to them are entities occurring in the AviationKG \cite{agarwal2022knowledge}. AviationQA will be helpful to researchers in finding insights into aircraft accidents and their prevention.
    % We contribute a large synthetic factoid QA dataset because that will be helpful to the Aviation industry.
    % how is it helpful? 
    
    \item We show that the size of a language model is inconsequential when knowledge is infused from the knowledge graphs. With AviationKG, we obtain 0.22, 0.23, and 0.23 hits@1 scores for link prediction using T5-small, T5-base, and T5-large, respectively. On AviationQA, we get a 0.70 hits@1 score on the three sizes of the T5 model. We validate our hypothesis with paired student t-test, and Cohen's kappa explained in section \ref{hypo-testing}. We obtain a substantial Cohen's kappa score of 0.76 for link prediction on AviationKG using T5-small and T5-large. For Question Answering using T5-small and T5-large, we get a Cohen's kappa score of 0.53 on the MetaQA dataset.   
    Hence, we provide evidence that we can substitute larger models with smaller ones and achieve the same performance with less computational cost and power. 
    
% We show that with the infusion of knowledge, the ability of the small and the large language model remains the same.    
\end{enumerate}

% sir's comments

% write how others can use our created QA pairs.
% it is a factoid dataset QA

\section{Motivation}
% write motivation
As stated earlier, in Section \ref{intro}, LMs are trained on generic datasets. So, knowledge from different sources, \textit{i.e.}, KGs, are used to perform downstream tasks in specific domain areas. LLMs infused with knowledge are required to perform such tasks, namely, QA and link prediction, which increases the need for computation power and time. We show that computational resources can be saved by using smaller language models for tasks.
\par
It is rare to obtain datasets related to the aviation domain, which is in increased demand.
We scrape NTSB reports from NTSB's website \footnote{\url{https://www.ntsb.gov/Pages/AviationQuery.aspx}} and create QA pairs that can be used by the aviation industry and researchers for Information Retrieval (IR) and QA purposes. The created dataset will help find insights into aircraft accidents and develop solutions to prevent accidents.

\section{ Background \& Related Work} \label{related work}
% Knowledge infusion using linked prediction KGT5 paper

% Link prediction in general, generally using KGE, cite KGEs
% Link Prediction using language models and why it works, cite KGT5

A Knowledge Graph is a collection of entities and relations represented in the form of triplets (subject, relation, object). Querying KG in Natural Language (NL) is a long-standing work. Early work focused on rule-based and pattern-based systems \cite{affolter2019comparative}. Recently, the work is shifted to seq2seq architecture \cite{zhong2017seq2sql} and pre-trained models with the advent of neural networks. Querying KGs remains a challenge because of the conversion of NL to the graph query language, namely, SPARQL, Cypher, etc.
\par
%  Improve next sentence
With the value increase of knowledge in the world, the popularity of the KG has escalated. Researchers are keenly interested in the synergy of knowledge graphs and deep learning. Several methods are exploited considering synergy: a) Integrating triplets of KG into the neural network \cite{Liu2020KBERTEL, saxena2022sequence}, b) Computing the relevance of entity and relations in a KG using a neural network \cite{sun-etal-2019-pullnet, yasunaga2021qa}.
\par 
Deep Learning models use representations of entities and relations to integrate triplets of KG. Knowledge Graph Embeddings are widely used to obtain representations \cite{KGESurveyelectronics9050750}. The KG embedding models are trained on link prediction over triplets to obtain representations \cite{LinkPredictionsym13030485}. Recent work has focused on using fine-tuned language models over KGE models for link prediction to reduce the number of parameters required to obtain the representations \cite{saxena2022sequence}.
\par
LMs and KGs are extensively used to improve task-specific performance. Still, no study has been done to understand the characteristics of a language model during the synergy of KG and DL.  
In this paper, we observe the behavior of language models after knowledge infusion with different domain datasets.  

% Can we change the title ofsection to Approach and Datasets
\section{Methodology and Experimental Design}
% Sir said me last time that for the main section description should not be written in a short paper
% In this section, we present our approach, experiment Data The experimental data first describes the data gathering and cleaning and then explains the model configurations and the evaluation technique.
This section presents our approach (flow diagram in figure \ref{fig:approach}), discusses the experiment datasets, creation of AviationQA, describes the model configurations, and explains the evaluation technique.

\subsection{Approach}
\label{approach}
\begin{figure*}[h]
  \centering
  \includegraphics[scale=0.20]{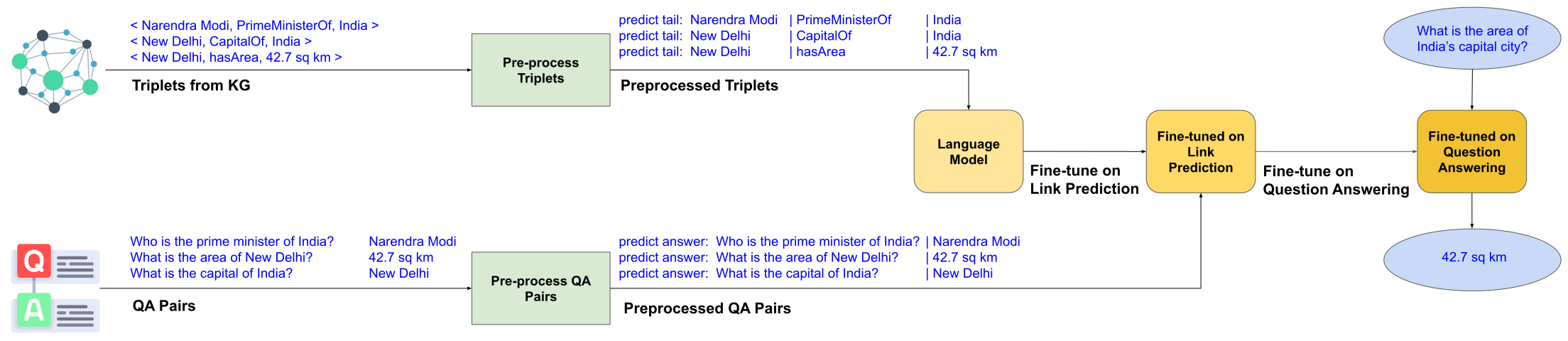}
  \caption{Flow diagram of the approach adopted in our paper. The model is first fine-tuned on KG triplets for Link Prediction. Next, the fine-tuned model is again fine-tuned on question answering. Because of the link-prediction task, the model learns KG completion and can answer multi-hop questions. E.g., If the model knows India's capital is New Delhi and New Delhi's area size, then the model should predict the area of India's capital correctly without explicitly mentioning New Delhi in the question}
  \label{fig:approach}
\end{figure*}

We observe the performance of small and large language models with the infusion of knowledge for link prediction and QA. Experiments are performed with the following models (detailed in section \ref{model_config}): a) T5-small non-pre-trained, b) T5-base pre-trained, c) T5-large pre-trained, and SOTA d) BLOOM 1b7.
We make use of different domain datasets for our approach, explained in section \ref{exp_data}. Figure \ref{fig:approach} demonstrates link prediction and question answering on the data after pre-processing.
\par
 We inject knowledge into the LMs. The knowledge is injected by the process of fine-tuning the pre-trained LM. Fine-tuning requires a learning objective and training data. In our case, the training data is triplets from the KG (table \ref{tab:triplets_stats}), and the learning objective is triple completion. Triple completion involves getting 
tail entity given head entity 
% entity-2 given entity-1 
and relation. Triple completion is also called link prediction. Thus, the LM absorbs the knowledge. The link prediction results with triplets are shown in table \ref{tab:link-prediction}.
\par
After fine-tuning on triplets for link prediction, the language model learns representations of entities and relations. The checkpoint with the best result on link prediction is used for the question-answering task. We again fine-tune the selected checkpoint with QA pairs (table \ref{tab:QA_stats}) and obtain the QA results shown in table \ref{tab:QA}.
% \par
% For the link prediction task, we begin by fine-tuning the language model over a knowledge base (table \ref{tab:triplets_stats}). The triplets are extracted from the knowledge base and structured in a format shown in section \ref{data pre-process} for fine-tuning.
% Then, we use fine-tuned model for predicting links on different datasets. Results for link prediction are shown in table \ref{tab:link-prediction}.
% \par
% After fine-tuning on triplets for link prediction, the language model learns representations of entities and relations. The checkpoint with the best result on link prediction is used for the question-answering task. We again fine-tune the selected checkpoint with QA pairs (table \ref{tab:QA_stats}) and obtain the QA results shown in table \ref{tab:QA}.
%%%%%%%%%%%%%%%%%%%%%%%%%
%%%%%%%%%%%%%%%%%%%%%%%%%
% domain specific KG are small, since these are small, larger models are not to gain much knowledge
% Even for larger KG, similar results with small and large langauage models
% Stengthens our intution in
% intution in Intor 
% Numbers in results
% Link prediction require very less parameters
% 

\subsection{Experiment Data} \label{exp_data}

We are using three datasets: a) Aviation Knowledge Graph (AviationKG) \cite{agarwal2022knowledge} \& Aviation QA pairs (section \ref{aviationQA}), b) MetaQA \cite{zhang2018variational}, which consists of a KB constructed from WikiMovies dataset \cite{miller-etal-2016-key} and question-answer pairs, and c) Complex Web Questions (CWQ) \cite{talmor2018web}, which uses subsets of Freebase \cite{chah2017freebase}. The statistic of these datasets is shown in table \ref{tab:triplets_stats} \& \ref{tab:QA_stats}.
We chose these datasets because they belong to different domains and vary in size. 
\par
MetaQA KB \& AviationKG are from the movie and aviation domains, respectively, which is useful to represent the diversity of datasets and validate our hypothesis. CWQ is based on Freebase, a huge KG, which is crowd-sourced.
We require a knowledge base and the corresponding QA pairs for our experimentation, described in section \ref{data_descp}.
MetaQA and CWQ are openly available datasets. But, there is no available QA pairs dataset for the aviation domain. We create a set of QA pairs in the aviation domain and contribute to the research community, detailed in section \ref{aviationQA}. The datasets used in the paper are pre-processed and split before running experiments, as explained in section \ref{data pre-process} and \ref{data_descp}.

\begin{table}[!h]
  \begin{center}
    \begin{tabular}{lrrr} 
    \hline
      \textbf{Dataset} & \textbf{Train} & \textbf{Validation} & \textbf{Test}\\
      \hline
      AviationKG & 173,372 & 10,000 & 10,000\\
      MovieKB & 249,482 & 10,000 & 10,000\\
      CWQ & 27,590,648 & 10,000 & 10,000\\
      \hline
    \end{tabular}
        \caption{Statistics of triplets (subject, relation, object) for three knowledge bases: AviationKG \cite{agarwal2022knowledge}, MetaKB \cite{zhang2018variational}, and Complex Web Question (CWQ) \cite{talmor2018web}. Subsets of Freebase \cite{chah2017freebase} are used for CWQ.}
    \label{tab:triplets_stats}
  \end{center}
\end{table}

\begin{table}[!h]
  \begin{center}
    \begin{tabular}{lrrr} 
    \hline
      \textbf{Dataset} & \textbf{Train} & \textbf{Validation} & \textbf{Test}\\
      \hline
      AviationQA & 367,304 & 10,000 & 10,000\\
      MetaQA  & 184,230 & 10,000 & 10,000\\
      CWQ & 61,619 & 3,519 & 3,531\\
      \hline
    \end{tabular}
        \caption{Statistics of Question Answer pairs from three domains: Aviation, Movie, and Web. For MetaQA, we use 1-hop questions. For more details, refer to section \ref{data_descp}. }
    \label{tab:QA_stats}
  \end{center}
\end{table}

\subsection{Data Pre-processing}\label{data pre-process}

We make use of KG and QA pairs (section \ref{exp_data}) from 3 domains, Aviation, Movie, and General domain. These datasets are cleaned and structured for our experiments. For the link prediction task, the dataset is created similar to \citet{saxena2022sequence}, described below:
\\
\textbf{predict head}: subject | relation | object
\\
\textbf{predict tail}: object | relation | subject
\\
The triplets \{subject, relation, object\} are extracted from the AviationKG, MovieKB, and Freebase individually. 
\par
All these knowledge bases are associated with the corresponding QA pairs. As explained in section \ref{aviationQA}, we construct the AviationQA pairs and use MetaQA 1-hop and CWQ for question answering. For QA fine-tuning, the dataset is in the given format:\\ \textbf{predict answer}: question |  answer. \\
E.g., \textbf{predict answer}: What is the capital of India? |  New Delhi.  
\\
Multiple answers exist for a question in AviationQA, MetaQA, and CWQ. These collective instances are separated as individual QA pairs. \\
E.g., What countries did Narendra Modi visit in the year 2021? Answers: United States, Italy. Every QA pair is segregated in the current layout: a) What countries did Narendra Modi visit in the year 2021? |  United States. b) What countries did Narendra Modi visit in the year 2021? |  Italy.
\par
With small KGs, \textit{i.e.}, AviationKG, and MovieKB, triplet samples are added during QA fine-tuning to avoid overfitting. The added triplets are in the same format as mentioned for the link prediction task.
The pre-processing of triplets and QA pairs is shown in figure \ref{fig:approach}.

\subsection{Creation of AviationQA}\label{aviationQA}

We web scrape the National Transportation Safety Board (NTSB) website and download 12k reports from 2009-2022. A set of 90 question templates is prepared using the common structure of documents in the format:
\begin{itemize}
    \item Where did the accident [ ] take place?
    
    \item What is the model/series of the aircraft bearing accident number [ ]? 
    
    \item Was there fire on the aircraft of the accident number [ ]?
\end{itemize}
 The template of questions is created, and answers to those questions are extracted from every NTSB report. Because every report is associated with an accident number, we place [ ] in the template to indicate which report the question pertains to, e.g., CHI07LA273, LAX07LA148.
NTSB reports are semi-structured, containing unstructured data in paragraphs and structured data in tabular format. We extract answers from each report w.r.t the template using the regular expression method.
 Later, QA pairs are scrutinized. As some reports' structure varies, different scripts are written to fetch answers for those reports. 
 % contribute to research community is redundant
 % scrutizine based on what?
\par
We successfully created 1 million factoid QA pairs in the aviation domain using the template-based method. The dataset will contribute to research and development in the aviation industry.

\subsection{Dataset Description}

\label{data_descp}

After pre-processing the data (section \ref{data pre-process}), we split it to train, validate, and test for link prediction and question answering. Table \ref{tab:triplets_stats} shows the split of triplets from AviationKG, MovieKB, and subsets of Freebase. CWQ uses subsets of Freebase, which is of size 27 million. AviationKG and MovieKB are domain-specific datasets of sizes 170k and 250k. Valid and test splits are equal in size to 10k each. 
\par
Our motive for considering different sizes and domain datasets is to strengthen our intuition that the performance of varying size models remains the same with an infusion of knowledge in language models. Table \ref{tab:link-prediction} shows the correctness of our intuition with the link prediction task.
% write abt why we split test and valid for 10k
\par
Table \ref{tab:QA_stats} shows the split of QA pairs for question-answering. We use 387,304 instances for AviationQA from 1 million QA pairs (section \ref{aviationQA}). The scrutinization is based on reports used to create AviationKG \cite{agarwal2022knowledge} from 1962 to 2015. We use QA pairs that have information available in the AviationKG. Moreover, we ensured that an answer to a question is an entity in the AviationKG.
\par
For comparison between the movie and the aviation data, the split of valid and test set is the same in both, \textit{i.e.}, 10k. CWQ dataset is smaller than AviationQA and MetaQA, so we use the same validation and test split, as mentioned in \citet{saxena2022sequence}.

% \subsubsection{Dataset Description} 

\subsection{Model Configuration} \label{model_config}
In this paper, we are using four models: T5-small non-pretrained (60 million parameters), T5-base pre-trained (220 million parameters), T5-large pre-trained (770 million parameters), and BLOOM (1.72 billion parameters).
% \cite{raffel2020exploring}.
These models are considered to validate our statement that with the injection of knowledge, small and large model performs the same. 
Both tasks, link prediction and question answering, are performed using these models.  
The T5 model is considered in our experiment as it is trained to perform multiple downstream tasks, i.e., translation, classification, and question answering. We use  BLOOM as it is similar to the SOTA model GPT-3 \cite{brown2020language}, which has outperformed other language models on tasks such as QA and summarization.

\subsection{Evaluation Technique}
\label{eval-tech}
We evaluate the performance of our models using the hits@1 score for link prediction and question answering. Table \ref{tab:link-prediction} and \ref{tab:QA} show the hits@1 score for link prediction and question answering, respectively, on different datasets. We choose the hits@1 score for evaluation as it is more precise than other hits@k scores. If the first predicted value matches the actual answer, then the score is 1; otherwise, 0. We are using the hits@1 metric and not other metrics such as BLEU score \cite{papineni2002bleu} and semantic similarity \cite{miller1991contextual} to validate the correctness of our hypothesis (introduced in section \ref{intro}). BLEU score is generally used for comparing sentences, whereas, for link prediction and QA tasks, the answer is a compound noun, \textit{i.e.}, an entity in the knowledge graph. Since the entities are ranked for tasks, the hits@1 score is the best metric. As the answers to link prediction and QA are entities of KG, the semantic similarity would not be able to distinguish between 2 different entities with semantically the same meaning. After considering all drawbacks of other metrics, we adapted the hits@1 score for the evaluation.

% Since the entities are considered distinct from each other, we don't use semantic similarity measures.

% Footnotes are inserted with the \verb|\footnote| command.\footnote{This is a footnote.}

\section{Results and Analysis}
\label{results}

\begin{table}[!]
  \begin{center}
    \begin{tabular}{lrrr} 
    \hline
      \textbf{Model} & \textbf{AviationKG} & \textbf{MetaKB} & \textbf{CWQ}\\
      \hline
      T5-small & 0.2258 & 0.0257 & 0.2153\\
      T5-base & 0.2387 & 0.0286 & 0.2273\\
      T5-large & 0.2323 & 0.0301 & 0.2207 \\
      BLOOM 1b7 & 0.2163 & 0.0365 & 0.2155 \\
      \hline
    \end{tabular}
        \caption{Link Prediction results on three knowledge bases: Aviation Knowledge Graph (KG) \cite{agarwal2022knowledge}, Meta Knowledge Base \cite{zhang2018variational}, and subsets of Freebase \cite{chah2017freebase} for Complex Web Questions (CWQ) \cite{talmor2018web}. }
    \label{tab:link-prediction}
  \end{center}
\end{table}

\begin{table}[!]
  \begin{center}
    \begin{tabular}{lrrr} 
    \hline
      \textbf{Model} & \textbf{AviationQA} & \textbf{MetaQA} & \textbf{CWQ}\\
      \hline
      T5-small & 0.7031 & 0.2144 & 0.2225\\
      T5-base & 0.7093 & 0.2158 & 0.2736\\
      T5-large & 0.7013 & 0.2371 & 0.2632 \\
       BLOOM 1b7 & 0.5507 & 0.2386 & 0.1517 \\
      \hline
    \end{tabular}
        \caption{Question Answering (QA) results in three QA datasets: AviationQA (\ref{aviationQA}), MetaQA \cite{zhang2018variational}, and Complex Web Questions (CWQ) \cite{talmor2018web}.}
    \label{tab:QA}
  \end{center}
\end{table}

\begin{table*}[!ht]

\begin{tabularx} {\columnwidth}{lccrrrr}

     \cline{1-7}
\multicolumn{1}{c}{\textbf{ \centering Hypothesis Testing}} & \multicolumn{3}{c}{ AviationKG} & \multicolumn{3}{c}{MetaQA} \\
\cmidrule(lr){2-4} \cmidrule(lr){5-7} 
&
\multicolumn{1}{p{1.6cm}}{\centering T5-small\\ T5-large} & \multicolumn{1}{p{1.6cm}}{\centering T5-base\\ T5-large} & \multicolumn{1}{p{1.6cm}}{\centering T5-large\\ Bloom} & \multicolumn{1}{p{1.6cm}}{\centering T5-small\\ T5-large} & \multicolumn{1}{p{1.6cm}}{\centering T5-base\\ T5-large} & \multicolumn{1}{p{1.6cm}}{\centering T5-large\\ Bloom} \\ 
     \cline{1-7}
      Paired Student T-test & \multicolumn{1}{p{1.6cm}}{\centering Cannot \\ Reject} & \multicolumn{1}{p{1.6cm}}{\centering Cannot \\ Reject} & \multicolumn{1}{p{1.6cm}}{\centering Cannot \\ Reject} & \multicolumn{1}{p{1.6cm}}{\centering Cannot \\ Reject} & \multicolumn{1}{p{1.6cm}}{\centering Cannot \\ Reject} & \multicolumn{1}{p{1.6cm}}{\centering Cannot \\ Reject} \\
      \cline{1-7}
   
Cohen's kappa Score & \multicolumn{1}{p{1.6cm}}{\centering 0.76} & \multicolumn{1}{p{1.6cm}}{\centering 0.75} & \multicolumn{1}{p{1.6cm}}{\centering 0.68} & \multicolumn{1}{p{1.6cm}}{\centering 0.49} & \multicolumn{1}{p{1.6cm}}{\centering 0.53} & \multicolumn{1}{p{1.6cm}}{\centering 0.33}  \\
    
Agreement (\%) & \multicolumn{1}{p{1.6cm}}{\centering 91.77} & \multicolumn{1}{p{1.6cm}}{\centering 91.36} & \multicolumn{1}{p{1.6cm}}{\centering 89.16} & \multicolumn{1}{p{1.6cm}}{\centering 82.50} & \multicolumn{1}{p{1.6cm}}{\centering 83.62} & \multicolumn{1}{p{1.6cm}}{\centering 75.73} \\
\cline{1-7}

\end{tabularx}
\caption{Hypothesis Testing on link prediction with `AviationKG' and question-answering with `MetaQA' datasets. We choose two measures for the test: a) paired Student T-test \cite{hsu2014paired}, and b) Cohen's kappa Score \cite{cohen1968weighted}, to prove our hypothesis- after injection of knowledge, small and large models perform the same. Student T-test with 0.1 significance value is done on 2000 instances of the test set selected randomly, and our hypothesis is not rejected 7 out of 10 times. We use the entire test set of 10,000 instances for the kappa score.
Cohen's kappa scores on link prediction for AviationKG are between 0.6 and 0.8, and on question-answering for MetaQA, between 0.4 and 0.6. With these scores, we are able to prove that our claim is correct. }
\label{hypotheis-test}
\end{table*}

This section analyzes the performance of two models: T5 and BLOOM. Table \ref{tab:link-prediction} \& \ref{tab:QA} show the hits@1 score for link prediction and QA tasks, respectively. With table \ref{tab:link-prediction}, we can clearly observe that the hits@1 score for three variations of the T5 model \& BLOOM is proximate for three different datasets (section \ref{data_descp}). The three T5 models score 0.22 \& 0.23 hits@1 for link prediction on AviationKG. Similarly, scores with MetaKB and CWQ have very less differences among models. LMs on MetaKB perform poorly for link prediction compared to other datasets; 0.02 \& 0.03 are the hits@1 scores on the T5 model \& BLOOM. The reason is the extensiveness of triplets in the MetaKB and the presence of noise in the dataset. We chose MetaKB to have a diversity of datasets and justify our claim (explained in section \ref{intro}).
%why performs poorly
\par
% CWQ scores 0.21 \& 0.22 for the T5 \& Bloom model on link prediction, consisting of 27 million triplets. 
The main observation with the link prediction task is that the T5-small non-pre-trained model performs alike to pre-trained models. The T5-base with 220 million parameters shows results like T5-large \& BLOOM, which comprises 770 million \& 1.7 billion parameters, respectively. 
Link prediction results (in table \ref{tab:link-prediction}) infers our claim that small and large models perform the same with the infusion of knowledge. 
\par
%write abt QA table.
To support our claim, we also performed QA with the same set of models as used for the link prediction task. With the AviationQA dataset, we achieved 0.7 hits@1 scores on T5-small, T5-base, and T5-large. LLMs such as T5-large \& BLOOM are expected to perform better for QA than small models as they are trained with a large amount of data and vice-versa, T5-small non-pre-trained, and T5-base are expected to perform direly. But, we observe that the performance of all three T5 models is the same for QA with the AviationQA dataset. Similarly, we observe that MetaQA achieves 0.2 hits@1 scores for non-pre-trained T5, pre-trained T5-base, T5-large, and BLOOM.
\par
% Research (section \ref{related work}) has shown that infusion of knowledge in language models is crucial for increasing the accuracy of a task and improving the performance of models. But, there has been no focus on how important is the size of a language model. In this paper, we have shown the requirement of the size of a language model for the downstream tasks, namely QA, after it learns the knowledge. Pre-trained and non-pre-trained models of different sizes have shown similar results on different domain datasets for link prediction and QA tasks. 
Through our experiments, we have shown how different model sizes perform on QA after infusion of knowledge using link prediction. Pre-trained and non-pre-trained models of different sizes have shown similar results on different domain datasets for link prediction and QA tasks. 
This contribution to the research community is pivotal as high accuracy can be achieved efficiently with less computation power, time, and cost.
\par
The source code for our paper is publicly available on GitHub\footnote{\url{https://github.com/ankush9812/Knowledge-Infusion-in-LM-for-QA}}. 

\section{Hypothesis Testing} 
\label{hypo-testing}

We attempt to contradict our hypothesis (\ref{intro}) that the difference in scores for the two models is negligible. We choose paired student t-test \cite{hsu2014paired} to refute our hypothesis. In our testing, the significance level (p-value) is 0.1, and the sample size is 20\% of the test set selected randomly. In comparing the pair of models (section \ref{model_config}), we predicted T5-large to perform better than T5-base \& T5-small and Bloom to perform better than all three models of T5 because of its large size. But, 7 out of 10 times student t-test was unable to reject our hypothesis, and the significance level among the pair of models was greater than 0.1. Table \ref{hypotheis-test} clearly shows the paired student t-test on AviationKG (table \ref{tab:triplets_stats}) and MetaQA (table \ref{tab:QA_stats}) for different pairs of models, and the result is the same, our hypothesis cannot be rejected. 
\par
After not being able to reject the hypothesis, our next step was to strengthen it, so, we calculate Cohen's kappa \cite{cohen1968weighted} score of the pair of models with different datasets (table \ref{tab:triplets_stats} \& \ref{tab:QA_stats}). We consider a pair of models as two annotators and the hits@1 score corresponding to each sample in the test set as their annotations. 
% Hits@1 score is the evaluation technique (\ref{eval-tech}) for both link prediction and question-answering.
Since our evaluation technique (section \ref{eval-tech}) uses hits@1 score and the score is binary for each sample, Cohen's kappa score is used to measure the reliability between the two models.
% In our paper, hits@1 score is the evaluation technique (\ref{eval-tech}) for both link prediction and question-answering. 
% So, a pair of models is assumed to be two annotators, and 
% Kappa score measures the reliability between them. 
The kappa score is calculated for all instances of the test set.
Table \ref{hypotheis-test} shows the Cohen's kappa score and \% agreement for AviationKG and MetaQA datasets between pair of models. For link prediction on AviationKG, the kappa score is between 0.6 and 0.8, and agreement is near 90\%. These results clearly denote the substantiality of our claim with high scores. We extend the test for question-answering with MetaQA. The pair of T5 models score 0.4-0.6, denoting moderate agreement as more than 80\% of agreement. T5-large and Bloom pair scores 0.33 with 75.7\% agreement, which is fair. 
\par
Thus, the testing supports our hypothesis, and we prove that the level of performance of different models with the infusion of knowledge remains the same.

\section{Conclusion and Future Work}
We have successfully created a million factoid QA pairs from the NTSB aircraft accident reports. The QA pairs are used in our experiments with AviationKG. 
We have validated our claim that with the infusion of knowledge to language models, the performance of the small language model is similar to the large language model. We substantiate with different language models and a diversity of datasets. Our investigation will benefit researchers in selecting the appropriate language model when working with knowledge and save computation power and time.
\par
The future line of work is to investigate the performance of models with incomplete and noisy knowledge graphs and study the extent to which the models can outright the domain knowledge. 
% complexQuestions and multi-hop questions

% \newpage

\section*{Acknowledgements}
This research is supported by the Science and Education Research Board (SERB), Ministry of Education, India, under the Imprint-2 project. We thank our Industry partner, Honeywell Technology Solutions Pvt Ltd, who provided insight and expertise that greatly assisted this research.

\nocite{*}
\bibliography{anthology}

\begin{thebibliography}{25}
\expandafter\ifx\csname natexlab\endcsname\relax\def\natexlab#1{#1}\fi

\bibitem[{Affolter et~al.(2019)Affolter, Stockinger, and
  Bernstein}]{affolter2019comparative}
Katrin Affolter, Kurt Stockinger, and Abraham Bernstein. 2019.
\newblock A comparative survey of recent natural language interfaces for
  databases.
\newblock \emph{The VLDB Journal}, 28(5):793--819.

\bibitem[{Agarwal et~al.(2022)Agarwal, Gite, Laddha, Bhattacharyya, Kar, Ekbal,
  Thind, Zele, and Shankar}]{agarwal2022knowledge}
Ankush Agarwal, Raj Gite, Shreya Laddha, Pushpak Bhattacharyya, Satyanarayan
  Kar, Asif Ekbal, Prabhjit Thind, Rajesh Zele, and Ravi Shankar. 2022.
\newblock Knowledge graph--deep learning: A case study in question answering in
  aviation safety domain.
\newblock \emph{arXiv preprint arXiv:2205.15952}.

\bibitem[{Brown et~al.(2020)Brown, Mann, Ryder, Subbiah, Kaplan, Dhariwal,
  Neelakantan, Shyam, Sastry, Askell et~al.}]{brown2020language}
Tom Brown, Benjamin Mann, Nick Ryder, Melanie Subbiah, Jared~D Kaplan, Prafulla
  Dhariwal, Arvind Neelakantan, Pranav Shyam, Girish Sastry, Amanda Askell,
  et~al. 2020.
\newblock Language models are few-shot learners.
\newblock \emph{Advances in neural information processing systems},
  33:1877--1901.

\bibitem[{Chah(2017)}]{chah2017freebase}
Niel Chah. 2017.
\newblock Freebase-triples: A methodology for processing the freebase data
  dumps.
\newblock \emph{arXiv preprint arXiv:1712.08707}.

\bibitem[{Cohen(1968)}]{cohen1968weighted}
Jacob Cohen. 1968.
\newblock Weighted kappa: nominal scale agreement provision for scaled
  disagreement or partial credit.
\newblock \emph{Psychological bulletin}, 70(4):213.

\bibitem[{Dai et~al.(2020)Dai, Wang, Xiong, and
  Guo}]{KGESurveyelectronics9050750}
Yuanfei Dai, Shiping Wang, Neal~N. Xiong, and Wenzhong Guo. 2020.
\newblock \href {https://doi.org/10.3390/electronics9050750} {A survey on
  knowledge graph embedding: Approaches, applications and benchmarks}.
\newblock \emph{Electronics}, 9(5).

\bibitem[{Hsu and Lachenbruch(2014)}]{hsu2014paired}
Henry Hsu and Peter~A Lachenbruch. 2014.
\newblock Paired t test.
\newblock \emph{Wiley StatsRef: statistics reference online}.

\bibitem[{Kale et~al.(2022)Kale, Bhattacharyya, Shetty, Gune, Shrivastava,
  Lawyer, and Biswas}]{kaveriradiology}
Kaveri Kale, Pushpak Bhattacharyya, Aditya Shetty, Milind Gune, Kush
  Shrivastava, Rustom Lawyer, and Spriha Biswas. 2022.
\newblock \href {https://doi.org/10.48550/ARXIV.2206.06308} {Knowledge graph
  construction and its application in automatic radiology report generation
  from radiologist's dictation}.

\bibitem[{Liu et~al.(2020)Liu, Zhou, Zhao, Wang, Ju, Deng, and
  Wang}]{Liu2020KBERTEL}
Weijie Liu, Peng Zhou, Zhe Zhao, Zhiruo Wang, Qi~Ju, Haotang Deng, and Ping
  Wang. 2020.
\newblock K-bert: Enabling language representation with knowledge graph.
\newblock In \emph{AAAI}.

\bibitem[{Michalopoulos et~al.(2021)Michalopoulos, Wang, Kaka, Chen, and
  Wong}]{michalopoulos-etal-2021-umlsbert}
George Michalopoulos, Yuanxin Wang, Hussam Kaka, Helen Chen, and Alexander
  Wong. 2021.
\newblock \href {https://doi.org/10.18653/v1/2021.naacl-main.139}
  {{U}mls{BERT}: Clinical domain knowledge augmentation of contextual
  embeddings using the {U}nified {M}edical {L}anguage {S}ystem
  {M}etathesaurus}.
\newblock In \emph{Proceedings of the 2021 Conference of the North American
  Chapter of the Association for Computational Linguistics: Human Language
  Technologies}, pages 1744--1753, Online. Association for Computational
  Linguistics.

\bibitem[{Miller et~al.(2016{\natexlab{a}})Miller, Fisch, Dodge, Karimi,
  Bordes, and Weston}]{kvmNet}
Alexander Miller, Adam Fisch, Jesse Dodge, Amir-Hossein Karimi, Antoine Bordes,
  and Jason Weston. 2016{\natexlab{a}}.
\newblock \href {https://doi.org/10.18653/v1/D16-1147} {Key-value memory
  networks for directly reading documents}.
\newblock In \emph{Proceedings of the 2016 Conference on Empirical Methods in
  Natural Language Processing}, pages 1400--1409, Austin, Texas. Association
  for Computational Linguistics.

\bibitem[{Miller et~al.(2016{\natexlab{b}})Miller, Fisch, Dodge, Karimi,
  Bordes, and Weston}]{miller-etal-2016-key}
Alexander Miller, Adam Fisch, Jesse Dodge, Amir-Hossein Karimi, Antoine Bordes,
  and Jason Weston. 2016{\natexlab{b}}.
\newblock \href {https://doi.org/10.18653/v1/D16-1147} {Key-value memory
  networks for directly reading documents}.
\newblock In \emph{Proceedings of the 2016 Conference on Empirical Methods in
  Natural Language Processing}, pages 1400--1409, Austin, Texas. Association
  for Computational Linguistics.

\bibitem[{Miller and Charles(1991)}]{miller1991contextual}
George~A Miller and Walter~G Charles. 1991.
\newblock Contextual correlates of semantic similarity.
\newblock \emph{Language and cognitive processes}, 6(1):1--28.

\bibitem[{Papineni et~al.(2002)Papineni, Roukos, Ward, and
  Zhu}]{papineni2002bleu}
Kishore Papineni, Salim Roukos, Todd Ward, and Wei-Jing Zhu. 2002.
\newblock Bleu: a method for automatic evaluation of machine translation.
\newblock In \emph{Proceedings of the 40th annual meeting of the Association
  for Computational Linguistics}, pages 311--318.

\bibitem[{Raffel et~al.(2020)Raffel, Shazeer, Roberts, Lee, Narang, Matena,
  Zhou, Li, Liu et~al.}]{raffel2020exploring}
Colin Raffel, Noam Shazeer, Adam Roberts, Katherine Lee, Sharan Narang, Michael
  Matena, Yanqi Zhou, Wei Li, Peter~J Liu, et~al. 2020.
\newblock Exploring the limits of transfer learning with a unified text-to-text
  transformer.
\newblock \emph{J. Mach. Learn. Res.}, 21(140):1--67.

\bibitem[{Roberts et~al.(2020)Roberts, Raffel, and Shazeer}]{roberts2020much}
Adam Roberts, Colin Raffel, and Noam Shazeer. 2020.
\newblock \href {https://doi.org/10.18653/v1/2020.emnlp-main.437} {How much
  knowledge can you pack into the parameters of a language model?}
\newblock In \emph{Proceedings of the 2020 Conference on Empirical Methods in
  Natural Language Processing (EMNLP)}, pages 5418--5426, Online. Association
  for Computational Linguistics.

\bibitem[{Saxena et~al.(2022)Saxena, Kochsiek, and
  Gemulla}]{saxena2022sequence}
Apoorv Saxena, Adrian Kochsiek, and Rainer Gemulla. 2022.
\newblock Sequence-to-sequence knowledge graph completion and question
  answering.
\newblock In \emph{Proceedings of the 60th Annual Meeting of the Association
  for Computational Linguistics (Volume 1: Long Papers)}, pages 2814--2828.

\bibitem[{Saxena et~al.(2020)Saxena, Tripathi, and Talukdar}]{embedkgqa}
Apoorv Saxena, Aditay Tripathi, and Partha Talukdar. 2020.
\newblock \href {https://doi.org/10.18653/v1/2020.acl-main.412} {Improving
  multi-hop question answering over knowledge graphs using knowledge base
  embeddings}.
\newblock In \emph{Proceedings of the 58th Annual Meeting of the Association
  for Computational Linguistics}, pages 4498--4507, Online. Association for
  Computational Linguistics.

\bibitem[{Sun et~al.(2019)Sun, Bedrax-Weiss, and Cohen}]{sun-etal-2019-pullnet}
Haitian Sun, Tania Bedrax-Weiss, and William Cohen. 2019.
\newblock \href {https://doi.org/10.18653/v1/D19-1242} {{P}ull{N}et: Open
  domain question answering with iterative retrieval on knowledge bases and
  text}.
\newblock In \emph{Proceedings of the 2019 Conference on Empirical Methods in
  Natural Language Processing and the 9th International Joint Conference on
  Natural Language Processing (EMNLP-IJCNLP)}, pages 2380--2390, Hong Kong,
  China. Association for Computational Linguistics.

\bibitem[{Talmor and Berant(2018)}]{talmor2018web}
Alon Talmor and Jonathan Berant. 2018.
\newblock \href {https://doi.org/10.18653/v1/N18-1059} {The web as a
  knowledge-base for answering complex questions}.
\newblock In \emph{Proceedings of the 2018 Conference of the North {A}merican
  Chapter of the Association for Computational Linguistics: Human Language
  Technologies, Volume 1 (Long Papers)}, pages 641--651, New Orleans,
  Louisiana. Association for Computational Linguistics.

\bibitem[{Wang et~al.(2021)Wang, Qiu, and Wang}]{LinkPredictionsym13030485}
Meihong Wang, Linling Qiu, and Xiaoli Wang. 2021.
\newblock \href {https://doi.org/10.3390/sym13030485} {A survey on knowledge
  graph embeddings for link prediction}.
\newblock \emph{Symmetry}, 13(3).

\bibitem[{Wang et~al.(2017)Wang, Mao, Wang, and Guo}]{wang2017knowledge}
Quan Wang, Zhendong Mao, Bin Wang, and Li~Guo. 2017.
\newblock Knowledge graph embedding: A survey of approaches and applications.
\newblock \emph{IEEE Transactions on Knowledge and Data Engineering},
  29(12):2724--2743.

\bibitem[{Yasunaga et~al.(2021)Yasunaga, Ren, Bosselut, Liang, and
  Leskovec}]{yasunaga2021qa}
Michihiro Yasunaga, Hongyu Ren, Antoine Bosselut, Percy Liang, and Jure
  Leskovec. 2021.
\newblock \href {https://doi.org/10.18653/v1/2021.naacl-main.45} {{QA}-{GNN}:
  Reasoning with language models and knowledge graphs for question answering}.
\newblock In \emph{Proceedings of the 2021 Conference of the North American
  Chapter of the Association for Computational Linguistics: Human Language
  Technologies}, pages 535--546, Online. Association for Computational
  Linguistics.

\bibitem[{Zhang et~al.(2018)Zhang, Dai, Kozareva, Smola, and
  Song}]{zhang2018variational}
Yuyu Zhang, Hanjun Dai, Zornitsa Kozareva, Alexander~J Smola, and Le~Song.
  2018.
\newblock Variational reasoning for question answering with knowledge graph.
\newblock In \emph{Thirty-second AAAI conference on artificial intelligence}.

\bibitem[{Zhong et~al.(2017)Zhong, Xiong, and Socher}]{zhong2017seq2sql}
Victor Zhong, Caiming Xiong, and Richard Socher. 2017.
\newblock Seq2sql: Generating structured queries from natural language using
  reinforcement learning.
\newblock \emph{arXiv preprint arXiv:1709.00103}.

\end{thebibliography}
\bibliographystyle{acl_natbib}

\appendix

\section{Appendix}
\label{sec:appendix}

% This is an appendix.
% \appendix

%Ques lrec paper reports<year 
% write why QA pairs split and number

\subsection{Examples of AviationQA}
Below, we mention some examples from our created Aviation question-answering dataset (section \ref{aviationQA}):
\begin{itemize}
    \item \textbf{Q:} Which seat was occupied by the pilot responsible for accident no. CEN18LA272?\\	\textbf{A:} Left
    
    \item \textbf{Q:} Are there other Aircraft Rating(s) for the pilot of accident no. GAA18CA489?\\
    \textbf{A:} None
    
    \item \textbf{Q:} What is the make of the aircraft bearing accident no. CEN18LA272?	\\
    \textbf{A:} Cessna
    
    \item \textbf{Q:} What is the category of the aircraft involved in accident no. GAA18CA489?\\	\textbf{A:} Gyroplane
    
    \item \textbf{Q:} What is the Airworthiness Certificate of accident no. GAA18CA297?\\	\textbf{A:} Normal
\end{itemize}

\end{document}